\documentclass[conference]{IEEEtran}
\IEEEoverridecommandlockouts
\usepackage{cite}
\usepackage{amsmath,amssymb,amsfonts}
\usepackage{algorithmic}
\usepackage{graphicx}
\usepackage{textcomp}
\usepackage{xcolor}
\def\BibTeX{{\rm B\kern-.05em{\sc i\kern-.025em b}\kern-.08em
		T\kern-.1667em\lower.7ex\hbox{E}\kern-.125emX}}
\begin{document}
	\bstctlcite{IEEEexample:BSTcontrol}
	
	\title{CNN Feature Map Augmentation for Single-Source Domain Generalization
		
		\thanks{The work leading to these results has been funded by the European Union under Grant Agreement No. 101057821, project RELEVIUM. Views and opinions expressed are however those of the authors and do not necessarily reflect those of the European Union or the granting authority (HaDEA). Neither the European Union nor the granting authority can be held responsible for them.}
	}
	
	\author{\IEEEauthorblockN{Aristotelis Ballas}
		\IEEEauthorblockA{\textit{Dept. of Informatics and Telematics} \\
			\textit{Harokopio University of Athens}\\
			Athens, Greece \\
			aballas@hua.gr}
		\and
		\IEEEauthorblockN{Christos Diou}
		\IEEEauthorblockA{\textit{Dept. of Informatics and Telematics} \\
			\textit{Harokopio University of Athens}\\
			Athens, Greece \\
			cdiou@hua.gr}
	}

	\maketitle

	\begin{abstract}
		In search of robust and generalizable machine learning models, Domain Generalization (DG) has gained significant traction during the past few years.
		The goal in DG is to produce models which continue to perform well when presented with data distributions different from the ones available during training. While deep convolutional neural networks (CNN) have been able to achieve outstanding performance on downstream computer vision tasks, they still often fail to generalize on previously \textit{unseen} data \textit{Domains}. Therefore, in this work we focus on producing a model which is able to remain robust under data distribution shift and propose an alternative regularization technique for convolutional neural network architectures in the single-source DG image classification setting. To mitigate the problem caused by domain shift between source and target data, we propose augmenting intermediate feature maps of CNNs. Specifically, we pass them through a novel \textit{Augmentation Layer} to prevent models from overfitting on the training set and improve their cross-domain generalization. To the best of our knowledge, this is the first paper proposing such a setup for the DG image classification setting. Experiments on the DG benchmark datasets of PACS, VLCS, Office-Home and TerraIncognita validate the effectiveness of our method, in which our model surpasses state-of-the-art algorithms in most cases.
	\end{abstract}
	

	\section{Introduction}
	Generalizing to unseen data is an ability inherit to humans yet quite challenging for machines to mimic. For example, it is quite apparent that an image of a dog running on grass, shares similar attributes with a dog lying on sand. As trivial as this may seem to us, machines often fail miserably when called upon to pinpoint these invariant attributes between images of the same class \cite{recht2019imagenet}. To this end, improving the generalization ability of Deep Learning (DL) models has been a long standing problem in the machine learning literature \cite{wang2022generalizing}. While advances have undoubtedly been made, in which models have even surpassed human experts in their respective fields \cite{he2015delving, mckinney_international_2020}, state-of-the-art algorithms often fail to maintain their accuracy when presented with previously `unseen' data.
	
	The vast majority of DL models are trained under the i.i.d assumption, meaning that their training (source) and test (target) data are identically and independently distributed. In practice however, source and target data seldom originate from similar data \textit{Domains} and are therefore out-of-distribution (OOD). To mitigate the problem caused by the domain shift present in such data, \textit{Domain Generalization} (DG) \cite{zhou2021domain} algorithms focus on producing models which retain their representation extraction capabilities when evaluated on OOD data. To validate their generalization ability and robustness, DG models are trained only on data originating from \textit{Source} domains and evaluated on data from, held-out during training, \textit{Target} domains. Most DG methods aim at pushing the model to learn \textit{invariant} \cite{arjovsky_invariant_2020} representations, which will enable them to remain unaffected by the distribution shift between data domains and avoid overfitting on the source data. Most common approaches include model regularization techniques, such as data augmentation or sample synthesis, adversarial training, bias simulation, meta-learning or learning via self-supervision (more in Section \ref{sec:related_work}).

	\begin{figure*}[t]
		\centering
		\includegraphics[width=\textwidth]{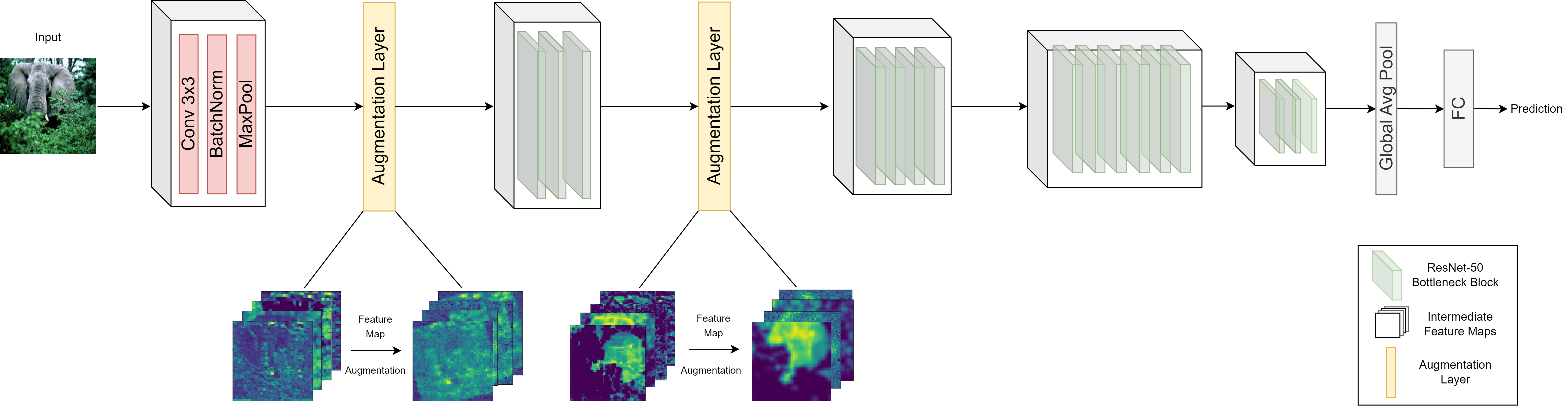}
		\caption{Visualization of the proposed method. The backbone of the architecture is a vanilla ResNet-50. We select to add our \textit{Augmentation Layer} after the first Max Pooling layer and after the third bottleneck block of the network. For the best performing implementation, a random 30\% of the feature maps are augmented in the first layer and 20\% in the second.}
		\label{model}
	\end{figure*}
	
	With regard to preventing overfitting, data augmentation techniques have proven exceptionally useful to deep convolutional neural networks and arguably to computer vision methods in general \cite{shorten_data_survey_2019}. Data augmentation techniques artificially amplify the training dataset by transforming input data, in order to generate additional latent space data points. The creation of additionally diversified datasets, enables classifiers to encapsulate the most significant aspects of the images and be less affected by domain shifts \cite{wang2017effectiveness}. On the other hand, regularization techniques aim at preventing overfitting by reducing the complexity of the model \cite{srivastava2014dropout}. %, ioffe_batch_2015}. 
Our proposed method lies somewhat in between data augmentation and model regularization. In this work, we argue that intermediate CNN feature map augmentation leads to model regularization and improved generalization ability in the DG image classification setting. To the best of our knowledge, this is the first paper to study the effects of CNN feature map augmentations in DG image classification. Specifically, in this work we:

\begin{itemize}
	\item Move past the classic approach of augmenting input data before feeding them to a deep CNN and research the affects of transforming a network's intermediate feature maps,
	\item Propose a simple and highly adaptable \textit{Augmentation layer} which applies transformations to such feature maps for improved model generalization and regularization,
	\item Demonstrate the robustness of our model by evaluating on the PACS 
	\cite{Li_2017_ICCV}, VLCS \cite{5995347}, Terra Incognita
	\cite{beery2018recognition} and Office-Home \cite{venkateswara2017deep} DG datasets and
	\item Provide an ablation study to illustrate the impact of each augmentation applied to the intermediate feature maps.
	%and explore promising combinations of CNN feature map augmentations for DG image classification. 
\end{itemize}

The following section presents some of the most significant papers proposed in the field of DG. In addition, we also discuss relevant DG methods which aim at model regularization and other works closest to our own. 

%Even with very few examples, we are able to capture the characteristics of a class which render it distinguishable and separate it from other classes or from the attributes that are insignificant for the task at hand. 

%For example in an image, an airplane has wings whether it is depicted on the ground or in the air, it is flying at night or at daytime etc, while an airplane is not a car since it has wings and turbines. 

%%%%%%%%%%%%%%%%%%%%%%%%%%%%%%%%%%%%%%%%%%%%%%%%%%%%%%%%%%%%%%%%%%%%%%%%%%%%%%%%
%%%%%%%%%%%%%%%%%%%%%%%%%%%%%%%%%%%%%%%%%%%%%%%%%%%%%%%%%%%%%%%%%%%%%%%%%%%%%%%%

\section{Related Work}
\label{sec:related_work}

\subsection{Domain Generalization}
When thinking of methods for addressing domain shift in data distributions, the most common approach would most likely be either Transfer Learning \cite{weiss2016survey} or Domain Adaptation\cite{wang2018deep}. 
While other algorithms do exist \cite{diou2010large}, DG methods differ in the fact that data from the target domain(s) remain \textit{unseen} throughout training. Therefore, the goal in DG is to be able to produce models which do not inference solely based on statistically correlated representations (e.g. due to selection biases in the dataset) but on causal attributes of the depicted class. In their survey, Zhou et al. \cite{zhou2022domainold} split DG algorithms into two categories: \textit{multi-source} and \textit{single-source}. The difference in the two categories is whether domain labels are known to the model during training. Multi-source DG algorithms focus on discovering invariances between the marginal distributions of separate, but known, data domains. On the contrary, in a much more difficult setting, single-source DG models have no information regarding the number or presence of distinct data domains and assume that their training data originates from a single distribution. Most papers in the literature propose multi-source methods, with a popular approach being Meta-Learning \cite{huisman2021survey}. For example, the models of \cite{li2018learning, pmlr-v70-finn17a} as well as Adaptive Risk Minimization \cite{zhang2021adaptive}, rely on previous experiment metadata for learning their optimal parameters. Furthermore, the authors of \cite{10.1007/978-3-030-58607-2_12} handle prediction uncertainty on unseen data domains by introducing a meta-learning probabilistic model which uses episodic training. In an another interesting approach, SagNets \cite{nam2021reducing} attempt to reduce style-biased predictions by using adversarial learning, to disentangle the style encodings in separate domains. Additionally, for minimizing the domain shift between source and target data distributions Deep Coral \cite{sun2016deep} aligns correlations of deep neural network layer activations, while MMD \cite{li2018domain} leverages the maximum mean discrepancy distance measure. Unlike multi-source methods, all single-source DG algorithms, such as our own, can be thought of as domain-agnostic. In their work \cite{ballasdiou2021}, the authors propose extracting multi-layer representations throughout a CNN and report improved performance against the baseline. Similarly, intermediate multi-layer representations have also been combined with self-attention mechanisms for image classification \cite{ballasdiouattention2023}, while they also seem to improve a model's robustness in the biosignal domain \cite{ballas2023towards}. To remove feature dependencies in their proposed model, the authors of \cite{Zhang_2021_CVPR} use sample weighting. Furthermore, \cite{kim_selfreg_2021} attempts to regularize the model into learning domain invariant features by using self-supervised losses. Finally, JiGen \cite{Carlucci_2019_CVPR} combines supervised and self-supervised learning, to solve jigsaw puzzles of its input images to learn the spatial correlation among image features.

\subsection{Data Augmentation \& Model Regularization}
Since our method lies between data augmentation and model regularization techniques, in this section we describe similar previous works. In the DG setting, the authors of \cite{huangRSC2020} introduce RSC, a training regularizing heuristic that improves the generalization ability of CNNs by discarding activations associated with high gradients. In \cite{zhou2021domain}, MixStyle augments its training data by using CNN feature statistics between images from different domains, while Mixup \cite{yan2020improve} mixes their pixel and feature space. Source domain input images have also been augmented with domain-adversarial gradients \cite{shankar2018generalizing, mansilla2021domain} but have been criticized as they are not able to simulate domain shifts found in real-world scenarios. Similarly, \cite{zhou2020learning} proposes augmenting training data by synthesizing pseudo-novel domains. Regarding CNN feature map augmentations, the number of previously proposed papers is limited. In \cite{kumar2020feature}, the authors introduce a Rotation Invariant Transformer (RiT) layer for improving rotation invariance in a small CNN network. In their model, the feature maps of the last convolutional layer are rotated using predefined variables, stacked, flattened and then passed through a fully connected layer for classification. Furthermore, \cite{kapoor_self-supervised_2020} proposes using a Feature Map Augmentation (FMA) loss for improving CNN robustness. However, the FMA loss is calculated between the representations of the original and augmented input images.

%%%%%%%%%%%%%%%%%%%%%%%%%%%%%%%%%%%%%%%%%%%%%%%%%%%%%%%%%%%%%%%%%%%%%%%%%%%%%%%%
%%%%%%%%%%%%%%%%%%%%%%%%%%%%%%%%%%%%%%%%%%%%%%%%%%%%%%%%%%%%%%%%%%%%%%%%%%%%%%%%

\section{Methodology}
\label{method}
\begin{figure}[t]
	\centering
	\includegraphics[width=0.98\columnwidth]{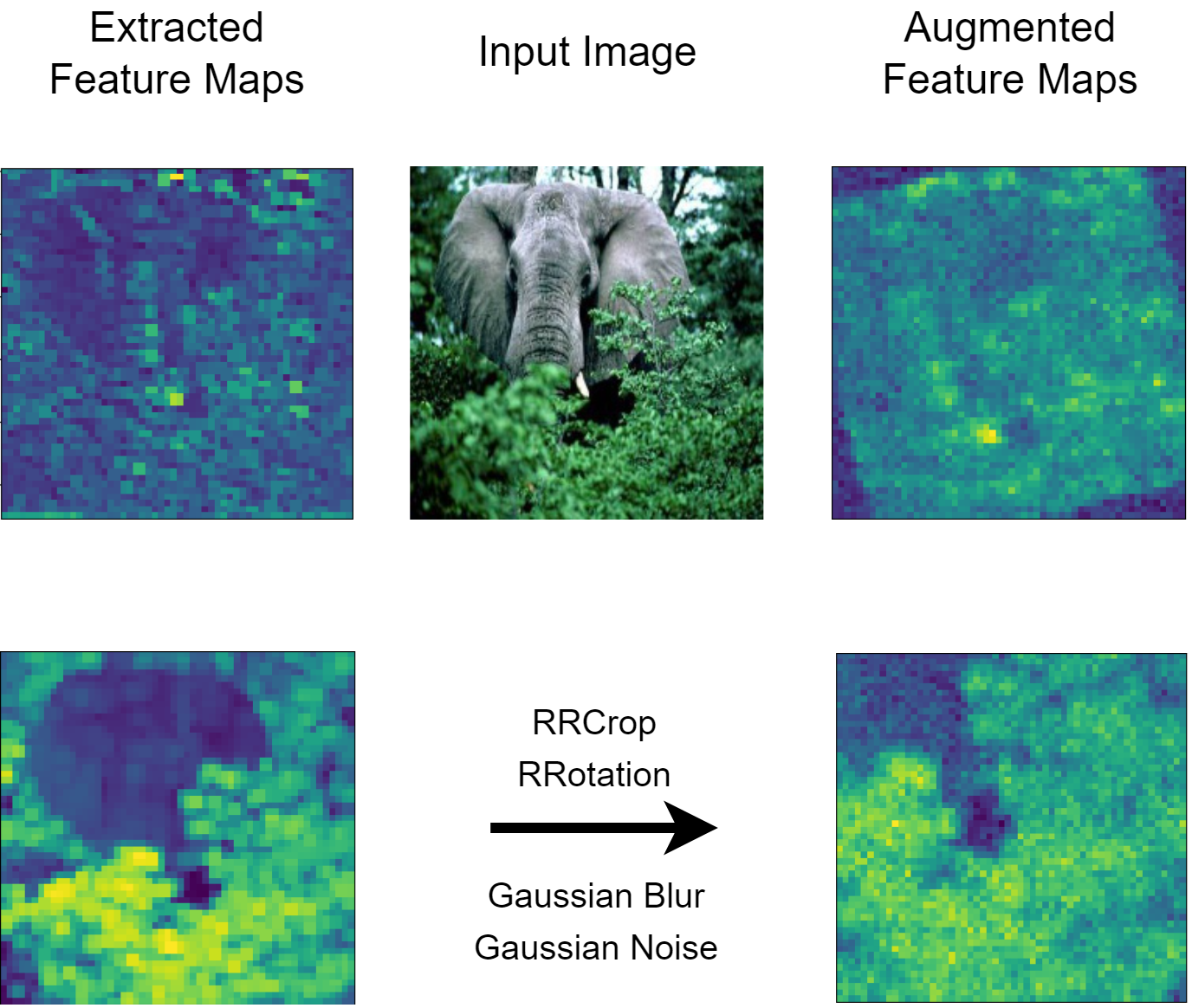}
	\caption{Visualization of the feature map transformations in our proposed \textit{Augmentation} layer, for the PACS \cite{Li_2017_ICCV} dataset. The images in the left column are feature maps extracted from the first Max Pooling layer of a vanilla ResNet-50 model. These feature maps are transformed via our proposed \textit{Augmentation Layer} before passing through the rest of the network. 
		%For the best performing implementation, a random 30\% of the feature maps are augmented. 
		The applied transformations are: Random Resized Crop, Random Rotation,, Random Horizontal flip, Gaussian Blur and Gaussian Noise addition. The result is a subset of transformed feature maps which act as regularizers and force the model to extract representations which remain invariant under such random perturbations.}
	\label{fmaps}
\end{figure}

Well-established CNN architectures, such as deep ResNet \cite{He_2016_CVPR} models, often fail to differentiate between causal and spurious correlations present in their training data distributions \cite{recht2019imagenet}. As such models tend to associate class labels with domain-dependent attributes (e.g background in images \cite{arjovsky_invariant_2020}), we claim that the transformation of the network's intermediate feature maps can act as a regularizer and push the model to learn domain invariant representations. 
For our proposed method, we introduce an \textit{Augmentation Layer} which transforms intermediate CNN feature maps before passing them through the rest of the backbone network. Specifically, let $\textbf{M} \in
\textbf{R}^{b \times c \times h \times w}$ be a batch of size ${b}$ intermediate CNN feature maps with ${c}$ channels, ${h}$ height and ${w}$ width each. Each feature map in the batch is passed through an Augmentation Layer ${A(\cdot, p)}$ which is a composition of image augmentation techniques. The ${p}$ parameter is a float between 0 and 1.0 and denotes the percentage of feature map channels that will be augmented. Therefore, the resulting perturbed feature maps are ${M' = A(M, p)}$, where ${p \times c}$ total channels have been transformed. In our implementation the Augmentation layer ${A}$ comprises of the following transformations: random resized crop, random horizontal flip, random rotation, Gaussian blur and addition of small Gaussian noise. The random resized crop function crops a channel of the feature map at a random location and then resizes the crop to reflect ${(h, w)}$. As its name suggests, the Random Horizontal Flip transformation horizontally flips the cropped map with a probability of 0.5. Similarly, the Random Rotation function randomly rotates the feature maps between 0 and 180 degrees, while Gaussian blur is also randomly added. Finally, we pollute the augmented maps with small noise drawn from a Gaussian distribution.. A visual example of the resulting transformed feature maps is provided in Figure \ref{fmaps}. We would also like to mention that during inference the input images are only passed through the layers of the regularized model without getting augmented by the Augmentation Layers.

Our proposed \textit{Augmentation Layer} is quite straightforward to implement and is highly adaptable. While an arbitrary number of transformations can be added to the augmentation composition in ${A}$, it can also be applied to the feature maps of any layer in a CNN network for model regularization. 

%%%%%%%%%%%%%%%%%%%%%%%%%%%%%%%%%%%%%%%%%%%%%%%%%%%%%%%%%%%%%%%%%%%%%%%%%%%%%%%%
%%%%%%%%%%%%%%%%%%%%%%%%%%%%%%%%%%%%%%%%%%%%%%%%%%%%%%%%%%%%%%%%%%%%%%%%%%%%%%%%

\section{Experimental Setup}
For our experiments, we select to implement our method on a ResNet-50 \cite{He_2016_CVPR} backbone network, pretrained on ImageNet and developed in PyTorch. The architecture of our model is presented in Figure \ref{model} where we select to add two Augmentation Layers: one after the network's first max pooling layer and one after its third bottleneck block. In our experiments, we found that the augmentation of 30\% random feature map channels (${p=0.3}$) in the first Augmentation Layer and 20\% in the second, leads to the best results. We train our model using the DomainBed \cite{gulrajani2021in} codebase, with an SGD optimizer, a learning rate of 0.001 and a batch-size of 128 images.  

We assess our method against the following previously proposed state-of-the-art DG algorithms: ERM\cite{vapnik1999nature}, RSC \cite{huangRSC2020}, MIXUP \cite{yan2020improve}, CORAL\cite{sun2016deep}, MMD \cite{li2018domain}, SagNet \cite{nam2021reducing}, SelfReg \cite{kim_selfreg_2021} and ARM \cite{zhang2021adaptive}. As the above algorithms are a mixture of both multi-source and single-source, the experimental results reflect the effectiveness of our model against methods which also take advantage of domain label knowledge. For a fair comparison, all baselines are implemented once again with DomainBed while their hyperparameters are set to reflect the values proposed in their original papers. Our method and all baselines are trained on a NVIDIA RTX A5000 GPU. Furthermore, the presented experimental results are averaged over 3 runs. Finally, we note that standard augmentations\footnote{Standard PyTorch augmentations include: RandomResizedCrop, RandomHorizontalFlip, ColorJitter, RandomGrayscale and Normalize. Definitions available at: https://pytorch.org/vision/stable/transforms.html} are applied on the training data splits before feeding them to each model for training.

\subsection{Datasets}
The robustness and generalization ability of our model are evaluated on four established DG benchmark datasets. Specifically, we train our model on: PACS \cite{Li_2017_ICCV}, VLCS \cite{5995347}, TerraIncognita \cite{beery2018recognition} and Office-Home \cite{venkateswara2017deep}. 

\begin{itemize}
	\item \textbf{PACS} consists of 9,991 images from four distinct domains, namely Photo, Art, Cartoon and Sketch. Each domain contains labels for 7 classes.
	\item \textbf{VLCS} contains 5 classes of 10,729 images from the PASCAL VOC, LabelMe, Caltech 101 and SUN09 domain datasets.
	\item \textbf{Office-Home} contains images from the domains of Art, Clipart, Product and Real-World. This dataset consists of 65 classes and 15,588 images in total.
	\item \textbf{TerraIncognita} is a real-world image dataset with 24,788 images of 10 different animals in the wild, taken from 4 different locations. 
\end{itemize}

%% Results
\begin{table}\centering
	\begin{center}
		\caption{Top-1\% accuracy results, averaged over 3 runs, on the pacs, vlcs, office-home (denoted as office) and terraincognita (denoted as terra) datasets. The top results are highlighted in \textbf{bold} while the second best are underlined.}
		\label{tab:results}
		\begin{tabular}{|c||cccc|c|}
			\hline\noalign{}%\smallskip}
		Method & PACS & VLCS & OFFICE & TERRA & Average \\
		%\noalign{\smallskip}
		\hline
		%\noalign{\smallskip}
		ERM    
		& 83.32 & 76.82 & 62.27 & 46.25 & 68.42 \\
		RSC    
		& 83.62 & 75.96	& 66.09	& 46.60	& 68.07 \\
		CORAL  
		& 83.57	& 76.97	& \underline{68.60}	& 48.05	& \underline{69.30} \\
		MIXUP   
		& 83.70 & \textbf{78.62} & 68.17 & 46.05 & 69.14 \\
		MMD    
		& 82.82	& 76.72 & 67.12	& 46.30	& 68.24 \\
		SagNet 
		& \underline{84.46} & 76.29 & 66.42	& \underline{48.60} & 68.94 \\
		SelfReg 
		& 84.16 & 75.46 & 65.73 & 47.00 & 68.09 \\
		ARM   
		& 83.78 & 76.38	& 63.49	& 45.50	& 67.29 \\
		\hline
		\textbf{Ours}
		& \textbf{85.14}& \underline{77.52} & \textbf{69.66} & \textbf{50.30} & \textbf{70.65} \\
		\hline
		
	\end{tabular}
\end{center}
\end{table}

\subsection{Results}
The experimental results are presented in Table \ref{tab:results}. Our seemingly simple regularization method proves to be highly effective in the DG image classification setting. Our model is able to surpass the previous state-of-the-art in three out of four benchmark datasets. Specifically, in PACS our algorithm exceeds the second-best model by 0.68\%, in Office-Home by 1.06\% and in TerraIncognita by 1.7\%. Although our model does not surpass the previous methods in VLCS, it is still able to rank second. Overall the results seem promising, as even with the application of standard image transformations our model is able to surpass the baselines by 1.25\% on average.

\subsection{Ablation Study}
To validate the selection of the augmentations implemented in our Augmentation Layer, we performed an ablation study on the Office-Home dataset. As mentioned in Section \ref{method}, our layer is a composition of the random resized crop, random horizontal flip, random rotation, Gaussian blur and Gaussian noise addition transformations. In our experiments we found that the random resized crop function had an overall positive effect and therefore elect to always keep it as an augmentation. Table \ref{tab:ablation} summarizes the results of the ablation study. It is apparent that the combination of all five augmentations leads to the best performing model. Other than that, it also seems as the addition of small Gaussian noise has a negative impact when the model is evaluated on the Clipart domain. As clipart is by far the most difficult domain in Office-Home, the additional regularization imposed by the noise may cause the model's performance to further deteriorate rather than improve its robustness. Moreover, the random rotation of the feature maps also proves to overall boost the networks performance across domains.

%% Ablation Study Table
\begin{table}\centering
\begin{center}
	\caption{Ablation study reporting the significance of each augmentation in our proposed Augmentation Layer. For the experiments we implement the same model as in Figure \ref{model} and adopt the OFFICE-HOME dataset. Best results are highlighted in \textbf{bold} while the second best are underlined. The random resized crop, random horizontal flip, random rotation, gaussian blur and gaussian noise addition transformations are denoted as RRC, RHF, RR, GB and GN respectively, while the Art, Clipart, Photo and Real-world domains in OFFICE-HOME are denoted as A, C, P and R.}
	\label{tab:ablation}
	\begin{tabular}{|ccccc||cccc||c|}
		\hline\noalign{\smallskip}
		RRC & RHF & RR & GB & GN
		& A & C & P & R & Avg\\
		\noalign{\smallskip}
		\hline
		\noalign{\smallskip}
		\checkmark & - &  -   & - & \checkmark
		& 63.28 & 50.46 & 73.39 & 77.03 &  66.04  \\
		\checkmark & - &     -   & - & -
		& 62.31 & 52.29 & \underline{75.87} & 77.84 & 67.08 \\
		\checkmark & - &     -   & \checkmark & \checkmark
		& 64.70 & 51.78 & 75.45 & 77.16 & 67.27 \\
		\checkmark & - & - & \checkmark & -
		& 63.74 & 53.15 & 75.53 & 78.65 & 67.76 \\
		\checkmark & - & \checkmark & - & \checkmark
		& 62.72 & 52.29 & 74.54 & 77.62 & 66.79 \\
		\checkmark & - & \checkmark & - & -
		& 61.43 & 52.11  &  73.70 & 77.10  & 66.08  \\
		\checkmark & - &    \checkmark  & \checkmark  & \checkmark
		& 65.51 & 52.77 & 74.97  & 78.16 & 67.85 \\
		\checkmark & - &     \checkmark   & \checkmark  & -
		& 64.57  & 53.37  & 73.50  & 76.01 & 66.86  \\
		\checkmark & \checkmark &     -   & - & \checkmark
		& 64.00 &  51.77 & 74.01  & 77.36  & 66.79  \\
		\checkmark & \checkmark & - & - & -
		& 63.54 & 53.26 & 74.07  & 76.01 & 66.72  \\
		\checkmark & \checkmark & - & \checkmark & \checkmark
		& 65.97 & 50.85 & 75.31 & \underline{78.41}  & 67.63  \\
		\checkmark & \checkmark & - & \checkmark & -
		& 65.77 &  \underline{54.43} &  74.47 &  77.65 & 68.08 \\
		\checkmark & \checkmark &\checkmark & - & \checkmark
		& 63.03 & 51.77  & 74.83 & 77.59 &  66.81 \\
		\checkmark & \checkmark & \checkmark & - & -
		& 63.33 & 53.66  &  73.39 & 77.07  &  {66.86} \\
		\checkmark & \checkmark & \checkmark  & \checkmark  & -
		& \underline{66.18} & {54.18} & {75.51} & {77.84} & \underline{68.43} \\
		
		\hline 
		\checkmark & \checkmark & \checkmark  & \checkmark & \checkmark
		& \textbf{67.20}    & \textbf{55.41} & \textbf{76.35} 
		& \textbf{79.67} & \textbf{69.66} \\

		\hline
	\end{tabular}
\end{center}
\end{table}

%%%%%%%%%%%%%%%%%%%%%%%%%%%%%%%%%%%%%%%%%%%%%%%%%%%%%%%%%%%%%%%%%%%%%%%%%%%%%%%%
%%%%%%%%%%%%%%%%%%%%%%%%%%%%%%%%%%%%%%%%%%%%%%%%%%%%%%%%%%%%%%%%%%%%%%%%%%%%%%%%

\section{Conclusions and Future Work}

In this paper we research an alternative method for regularizing CNN models, in the single-source DG image classification setting. We argue that the augmentation of intermediate CNN feature maps can assist the model into extracting class-relevant and domain invariant representations from input images and improve its robustness against previously unseen data. %What's more, the above technique aids the model in avoiding overfitting on its training data
The effectiveness of our algorithm is supported by the experimental results on 4 widely adopted DG image classification benchmarks, in which it is able to surpass the baselines in three out of four cases. Regardless, our method has some drawbacks. Depending on the position of the Augmentation Layer in the network, it may have a negative effect in the model's performance, while the augmentations applied in the layers may not be appropriate for certain datasets. To that end, in a future work we intend to experiment with the positioning of our proposed Augmentation Layer throughout various backbone networks and report on the best practices for several benchmark datasets. Furthermore, we also aim to extend the idea of augmenting internal neural network distributions, such as CNN filters, develop additional transformation techniques, possibly combine them with attention mechanisms and finally explore the value of feature map augmentation on additional computer vision settings.

%\begin{table}[hp]
%\caption{Table Type Styles}
%\begin{center}
%\begin{tabular}{|c|c|c|c|}
%\hline
%\textbf{Table}&\multicolumn{3}{|c|}{\textbf{Table Column Head}} \\
%\cline{2-4} 
%\textbf{Head} & \textbf{\textit{Table column subhead}}& \textbf{\textit{Subhead}}& \textbf{\textit{Subhead}} \\
%\hline
%copy& More table copy$^{\mathrm{a}}$& &  \\
%\hline
%\multicolumn{4}{l}{$^{\mathrm{a}}$Sample of a Table footnote.}
%\end{tabular}
%\label{tab1}
%\end{center}
%\end{table}

%\begin{figure}[htbp]
%\centerline{\includegraphics{fig1.png}}
%\caption{Example of a figure caption.}
%\label{fig}
%\end{figure}

%\section*{Acknowledgment}

%The preferred spelling of the word ``acknowledgment'' in America is without 
%an ``e'' after the ``g''. Avoid the stilted expression ``one of us (R. B. 
%G.) thanks $\ldots$''. Instead, try ``R. B. G. thanks$\ldots$''. Put sponsor 
%acknowledgments in the unnumbered footnote on the first page.

\bibliographystyle{IEEEtran}
\bibliography{references}

% Generated by IEEEtran.bst, version: 1.14 (2015/08/26)
\begin{thebibliography}{10}
\providecommand{\url}[1]{#1}
\csname url@samestyle\endcsname
\providecommand{\newblock}{\relax}
\providecommand{\bibinfo}[2]{#2}
\providecommand{\BIBentrySTDinterwordspacing}{\spaceskip=0pt\relax}
\providecommand{\BIBentryALTinterwordstretchfactor}{4}
\providecommand{\BIBentryALTinterwordspacing}{\spaceskip=\fontdimen2\font plus
\BIBentryALTinterwordstretchfactor\fontdimen3\font minus
  \fontdimen4\font\relax}
\providecommand{\BIBforeignlanguage}[2]{{%
\expandafter\ifx\csname l@#1\endcsname\relax
\typeout{** WARNING: IEEEtran.bst: No hyphenation pattern has been}%
\typeout{** loaded for the language `#1'. Using the pattern for}%
\typeout{** the default language instead.}%
\else
\language=\csname l@#1\endcsname
\fi
#2}}
\providecommand{\BIBdecl}{\relax}
\BIBdecl

\bibitem{recht2019imagenet}
B.~Recht, R.~Roelofs, L.~Schmidt, and V.~Shankar, ``Do imagenet classifiers
  generalize to imagenet?'' in \emph{International Conference on Machine
  Learning}.\hskip 1em plus 0.5em minus 0.4em\relax PMLR, 2019, pp. 5389--5400.

\bibitem{wang2022generalizing}
J.~Wang, C.~Lan \emph{et~al.}, ``Generalizing to unseen domains: A survey on
  domain generalization,'' \emph{IEEE Transactions on Knowledge and Data
  Engineering}, 2022.

\bibitem{he2015delving}
K.~He, X.~Zhang, S.~Ren, and J.~Sun, ``Delving deep into rectifiers: Surpassing
  human-level performance on imagenet classification,'' in \emph{Proceedings of
  the IEEE international conference on computer vision}, 2015, pp. 1026--1034.

\bibitem{mckinney_international_2020}
S.~M. McKinney, M.~Sieniek \emph{et~al.},
  ``\BIBforeignlanguage{en}{International evaluation of an {AI} system for
  breast cancer screening},'' \emph{\BIBforeignlanguage{en}{Nature}}, vol. 577,
  no. 7788, pp. 89--94, Jan. 2020, number: 7788 Publisher: Nature Publishing
  Group.

\bibitem{zhou2021domain}
K.~Zhou, Y.~Yang, Y.~Qiao, and T.~Xiang, ``Domain generalization with
  mixstyle,'' in \emph{International Conference on Learning Representations},
  2021.

\bibitem{arjovsky_invariant_2020}
M.~Arjovsky, L.~Bottou, I.~Gulrajani, and D.~Lopez-Paz, ``Invariant {Risk}
  {Minimization},'' \emph{arXiv:1907.02893 [cs, stat]}, Mar. 2020, arXiv:
  1907.02893.

\bibitem{shorten_data_survey_2019}
C.~Shorten and T.~M. Khoshgoftaar, ``A survey on {Image} {Data} {Augmentation}
  for {Deep} {Learning},'' \emph{Journal of Big Data}, vol.~6, no.~1, p.~60,
  Jul. 2019.

\bibitem{wang2017effectiveness}
J.~Wang, L.~Perez \emph{et~al.}, ``The effectiveness of data augmentation in
  image classification using deep learning,'' \emph{Convolutional Neural
  Networks Vis. Recognit}, vol.~11, no. 2017, pp. 1--8, 2017.

\bibitem{srivastava2014dropout}
N.~Srivastava, G.~Hinton, A.~Krizhevsky, I.~Sutskever, and R.~Salakhutdinov,
  ``Dropout: a simple way to prevent neural networks from overfitting,''
  \emph{The journal of machine learning research}, vol.~15, no.~1, pp.
  1929--1958, 2014.

\bibitem{Li_2017_ICCV}
D.~Li, Y.~Yang, Y.-Z. Song, and T.~M. Hospedales, ``Deeper, broader and artier
  domain generalization,'' in \emph{Proceedings of the IEEE International
  Conference on Computer Vision (ICCV)}, 2017.

\bibitem{5995347}
A.~Torralba and A.~A. Efros, ``Unbiased look at dataset bias,'' in \emph{CVPR
  2011}, 2011.

\bibitem{beery2018recognition}
S.~Beery, G.~Van~Horn, and P.~Perona, ``Recognition in terra incognita,'' in
  \emph{Proceedings of the European conference on computer vision (ECCV)},
  2018, pp. 456--473.

\bibitem{venkateswara2017deep}
H.~Venkateswara, J.~Eusebio, S.~Chakraborty, and S.~Panchanathan, ``Deep
  hashing network for unsupervised domain adaptation,'' in \emph{Proceedings of
  the IEEE conference on computer vision and pattern recognition}, 2017, pp.
  5018--5027.

\bibitem{weiss2016survey}
K.~Weiss, T.~M. Khoshgoftaar, and D.~Wang, ``A survey of transfer learning,''
  \emph{Journal of Big data}, vol.~3, no.~1, pp. 1--40, 2016.

\bibitem{wang2018deep}
M.~Wang and W.~Deng, ``Deep visual domain adaptation: A survey,''
  \emph{Neurocomputing}, vol. 312, pp. 135--153, 2018.

\bibitem{diou2010large}
C.~Diou, G.~Stephanopoulos, P.~Panagiotopoulos, C.~Papachristou, N.~Dimitriou,
  and A.~Delopoulos, ``Large-scale concept detection in multimedia data using
  small training sets and cross-domain concept fusion,'' \emph{IEEE
  transactions on circuits and systems for video technology}, vol.~20, no.~12,
  pp. 1808--1821, 2010.

\bibitem{zhou2022domainold}
K.~Zhou, Z.~Liu, Y.~Qiao, T.~Xiang, and C.~C. Loy, ``Domain generalization: A
  survey,'' \emph{IEEE Transactions on Pattern Analysis and Machine
  Intelligence}, 2022.

\bibitem{huisman2021survey}
M.~Huisman, J.~N. Van~Rijn, and A.~Plaat, ``A survey of deep meta-learning,''
  \emph{Artificial Intelligence Review}, vol.~54, no.~6, pp. 4483--4541, 2021.

\bibitem{li2018learning}
D.~Li, Y.~Yang, Y.-Z. Song, and T.~M. Hospedales, ``Learning to generalize:
  Meta-learning for domain generalization,'' in \emph{Thirty-Second AAAI
  Conference on Artificial Intelligence}, 2018.

\bibitem{pmlr-v70-finn17a}
C.~Finn, P.~Abbeel, and S.~Levine, ``Model-agnostic meta-learning for fast
  adaptation of deep networks,'' in \emph{Proceedings of the 34th International
  Conference on Machine Learning}, ser. Proceedings of Machine Learning
  Research.\hskip 1em plus 0.5em minus 0.4em\relax PMLR, 2017.

\bibitem{zhang2021adaptive}
M.~Zhang, H.~Marklund, N.~Dhawan, A.~Gupta, S.~Levine, and C.~Finn, ``Adaptive
  risk minimization: Learning to adapt to domain shift,'' \emph{Advances in
  Neural Information Processing Systems}, vol.~34, pp. 23\,664--23\,678, 2021.

\bibitem{10.1007/978-3-030-58607-2_12}
Y.~Du, J.~Xu, H.~Xiong, Q.~Qiu, X.~Zhen, C.~G.~M. Snoek, and L.~Shao,
  ``Learning to learn with variational information bottleneck for domain
  generalization,'' in \emph{Computer Vision -- ECCV 2020}.\hskip 1em plus
  0.5em minus 0.4em\relax Cham: Springer International Publishing, 2020.

\bibitem{nam2021reducing}
H.~Nam, H.~Lee, J.~Park, W.~Yoon, and D.~Yoo, ``Reducing domain gap by reducing
  style bias,'' in \emph{Proceedings of the IEEE/CVF Conference on Computer
  Vision and Pattern Recognition}, 2021, pp. 8690--8699.

\bibitem{sun2016deep}
B.~Sun and K.~Saenko, ``Deep coral: Correlation alignment for deep domain
  adaptation,'' in \emph{European conference on computer vision}.\hskip 1em
  plus 0.5em minus 0.4em\relax Springer, 2016, pp. 443--450.

\bibitem{li2018domain}
H.~Li, S.~J. Pan, S.~Wang, and A.~C. Kot, ``Domain generalization with
  adversarial feature learning,'' in \emph{Proceedings of the IEEE conference
  on computer vision and pattern recognition}, 2018, pp. 5400--5409.

\bibitem{ballasdiou2021}
A.~Ballas and C.~Diou, ``Multi-layer representation learning for robust ood
  image classification,'' in \emph{Proceedings of the 12th Hellenic Conference
  on Artificial Intelligence}, ser. SETN '22.\hskip 1em plus 0.5em minus
  0.4em\relax New York, NY, USA: Association for Computing Machinery, 2022.

\bibitem{ballasdiouattention2023}
------, ``Cnns with multi-level attention for domain generalization,'' ser.
  ICMR '23.\hskip 1em plus 0.5em minus 0.4em\relax New York, NY, USA:
  Association for Computing Machinery, 2023.

\bibitem{ballas2023towards}
------, ``Towards domain generalization for ecg and eeg classification:
  Algorithms and benchmarks,'' \emph{arXiv preprint arXiv:2303.11338}, 2023.

\bibitem{Zhang_2021_CVPR}
X.~Zhang, P.~Cui, R.~Xu, L.~Zhou, Y.~He, and Z.~Shen, ``Deep stable learning
  for out-of-distribution generalization,'' in \emph{Proceedings of the
  IEEE/CVF Conference on Computer Vision and Pattern Recognition (CVPR)}, 2021.

\bibitem{kim_selfreg_2021}
D.~Kim, Y.~Yoo, S.~Park, J.~Kim, and J.~Lee,
  ``\BIBforeignlanguage{en}{{SelfReg}: {Self}-supervised {Contrastive}
  {Regularization} for {Domain} {Generalization}},'' in
  \emph{\BIBforeignlanguage{en}{2021 {IEEE}/{CVF} {International} {Conference}
  on {Computer} {Vision} ({ICCV})}}.\hskip 1em plus 0.5em minus 0.4em\relax
  Montreal, QC, Canada: IEEE, Oct. 2021, pp. 9599--9608.

\bibitem{Carlucci_2019_CVPR}
F.~M. Carlucci, A.~D'Innocente, S.~Bucci, B.~Caputo, and T.~Tommasi, ``Domain
  generalization by solving jigsaw puzzles,'' in \emph{Proceedings of the
  IEEE/CVF Conference on Computer Vision and Pattern Recognition (CVPR)}, 2019.

\bibitem{huangRSC2020}
Z.~Huang, H.~Wang, E.~P. Xing, and D.~Huang, ``Self-challenging improves
  cross-domain generalization,'' in \emph{ECCV}, 2020.

\bibitem{yan2020improve}
S.~Yan, H.~Song, N.~Li, L.~Zou, and L.~Ren, ``Improve unsupervised domain
  adaptation with mixup training,'' \emph{arXiv preprint arXiv:2001.00677},
  2020.

\bibitem{shankar2018generalizing}
S.~Shankar, V.~Piratla, S.~Chakrabarti, S.~Chaudhuri, P.~Jyothi, and
  S.~Sarawagi, ``Generalizing across domains via cross-gradient training,''
  \emph{arXiv preprint arXiv:1804.10745}, 2018.

\bibitem{mansilla2021domain}
L.~Mansilla, R.~Echeveste, D.~H. Milone, and E.~Ferrante, ``Domain
  generalization via gradient surgery,'' in \emph{Proceedings of the IEEE/CVF
  International Conference on Computer Vision}, 2021, pp. 6630--6638.

\bibitem{zhou2020learning}
K.~Zhou, Y.~Yang, T.~Hospedales, and T.~Xiang, ``Learning to generate novel
  domains for domain generalization,'' in \emph{Computer Vision--ECCV 2020:
  16th European Conference, Glasgow, UK, August 23--28, 2020, Proceedings, Part
  XVI 16}.\hskip 1em plus 0.5em minus 0.4em\relax Springer, 2020, pp. 561--578.

\bibitem{kumar2020feature}
D.~Kumar, D.~Sharma, and R.~Goecke, ``Feature map augmentation to improve
  rotation invariance in convolutional neural networks,'' in \emph{Advanced
  Concepts for Intelligent Vision Systems: 20th International Conference, ACIVS
  2020, Auckland, New Zealand, February 10--14, 2020, Proceedings 20}.\hskip
  1em plus 0.5em minus 0.4em\relax Springer, 2020, pp. 348--359.

\bibitem{kapoor_self-supervised_2020}
N.~Kapoor, C.~Yuan \emph{et~al.}, ``A {Self}-{Supervised} {Feature} {Map}
  {Augmentation} ({FMA}) {Loss} and {Combined} {Augmentations} {Finetuning} to
  {Efficiently} {Improve} the {Robustness} of {CNNs},'' in \emph{Proceedings of
  the 4th {ACM} {Computer} {Science} in {Cars} {Symposium}}, ser. {CSCS}
  '20.\hskip 1em plus 0.5em minus 0.4em\relax New York, NY, USA: Association
  for Computing Machinery, Dec. 2020, pp. 1--8.

\bibitem{He_2016_CVPR}
K.~He, X.~Zhang, S.~Ren, and J.~Sun, ``Deep residual learning for image
  recognition,'' in \emph{Proceedings of the IEEE Conference on Computer Vision
  and Pattern Recognition (CVPR)}, June 2016.

\bibitem{gulrajani2021in}
I.~Gulrajani and D.~Lopez-Paz, ``In search of lost domain generalization,'' in
  \emph{International Conference on Learning Representations}, 2021.

\bibitem{vapnik1999nature}
V.~Vapnik, \emph{The nature of statistical learning theory}.\hskip 1em plus
  0.5em minus 0.4em\relax Springer science \& business media, 1999.

\end{thebibliography}

%\vspace{12pt}

\end{document}